\documentclass{article}

\usepackage[utf8]{inputenc} 
\usepackage[T1]{fontenc}    
\usepackage{hyperref}       
\usepackage{url}            
\usepackage{booktabs}       
\usepackage{amsfonts}       
\usepackage{nicefrac}       
\usepackage{microtype}      
\usepackage{xcolor}         
\usepackage{pgfplotstable}

\usepackage{algorithm}
\usepackage{algorithmic}
\title{Are LLM-based methods good enough for detecting unfair terms of service?}
\author{  Mirgita Frasheri, Arian Bakhtiarnia, Lukas Esterle, Alexandros Iosifidis\\
  DIGIT, Department of Electrical and Computer Engineering\\
  Aarhus University\\
  Denmark \\
  \texttt{\{mirgita.frasheri, arianbakh, lukas.esterle, ai\}@ece.au.dk} 
  }

\begin{document}

\maketitle

\begin{abstract}
  Countless terms of service (ToS) are being signed everyday by users all over the world while interacting with all kinds of apps and websites. 
    More often than not, these online contracts spanning double-digit pages are signed blindly by users who simply want immediate access to the desired service.
    What would normally require a consultation with a legal team, has now become a mundane activity consisting of a few clicks where users potentially sign away their rights, for instance in terms of their data privacy, to countless online entities/companies.
    Large language models (LLMs) are good at parsing long text-based documents, and could potentially be adopted to help users when dealing with dubious clauses in ToS and their underlying privacy policies.
    To investigate the utility of existing models for this task, we first build a dataset consisting of 12 questions applied individually to a set of privacy policies crawled from popular websites. 
    Thereafter, a series of open-source as well as commercial chatbots such as ChatGPT, are queried over each question, with the answers being compared to a given ground truth. 
    Our results show that some open-source models are able to provide a higher accuracy compared to some commercial models. However, the best performance is recorded from a commercial chatbot (ChatGPT4). Overall, all models perform only slightly better than random at this task.
    Consequently, their performance needs to be significantly improved before they can be adopted at large for this purpose.
\end{abstract}

\section{Introduction}
Questionable contracts are being signed everyday by users all over the world while they are visiting or interacting with different websites and apps. 
Most users provide their signature almost instantly to documents spanning multiple pages, without reading through properly all the details and potentially unfair clauses contained in them.
In doing so they can easily sign away their rights to privacy to all types of online entities/companies.
Even seemingly innocuous contracts can contain clauses that are dubious. 
For example, the privacy policies of some gyms with online registration, which contain an invasive clause that empowers a staff member to request at any time urine samples from any gym goer to perform doping tests.
In case of refusal to comply, the test is considered positive and the `offending' individual could be banned or penalized in some way.

It has been argued that these contracts, also referred to as click-through or click-to-contract, are in fact only giving the illusion of consent to users~\footnote{\url{https://bit.ly/44tfce6}}.
They are devised from the get-go to be convoluted.
Vardi argues in the Communications of the ACM (vol. 66, 09)~\cite{vardi2023} that regulation is needed to nullify such contracts, shaping technology such that it is human-centered, and hindering what can be considered as predatory tech.  
Either agreeing or not that such regulation should take place, users around the world need concrete tools to deal with the large amount of click-through contracts appearing at every virtual corner. 

To tackle the problem of dealing with click-through contracts, focusing on privacy policies, we investigate whether large language models (LLMs) can be adopted to help users figure out whether there are unfair or dubious claims in click-through contracts.
Specifically, we contribute with a dataset consisting of 12 questions applied individually to a set of privacy policies from popular sites.
Thereafter, we query a series of open-source and commercial chatbots, and compare the results to the given ground-truth. 
Our dataset and code for replicating our experiments is publicly available online~\footnote{The link to the repository has been omitted due to the double-blind submission.}. 
Our results show that overall ChatGPT4 has the best performance for the task at hand, however, the best performing open-source chatbot can outperform ChatGPT3.
All chatbots, both open-source and commercial, generally perform better than a random strategy.

The rest of this paper is organized as follows. 
Related work is presented in Section~\ref{sec:related}. Section~\ref{sec:dataset} describes the dataset, and provides a comparison with related existing datasets. 
Methods for answering questions and text summarization are described (Section~\ref{sec:method}).
The experimental setup, followed by results are provided in Section~\ref{sec:experiments}, with the paper concluding in Section~\ref{sec:conclusion}.

\section{Related Work}\label{sec:related} 

Several works regarding the use of natural language processing (NLP) for legal purposes exist in the literature. Many methods focus on predicting the outcome, that is, the final judgment of legal cases~\cite{cui2023survey}. Korreda et al.~\cite{koreeda-manning-2021-contractnli-dataset} proposes a method for determining whether or not a given hypothesis is entailed by a contract. 

However, methods focusing on terms of service and privacy policy are rather limited. Lippi et al.~\cite{Lippi2019} formulates the problem of detecting unfair terms of service as a classification task, where each clause in a terms of service document is classified as either fair or unfair, where seven classes of unfairness exist. The issue with this formulation is that long terms of service documents first need to be broken down into short clauses, which is not practical for end-users. Furthermore, fairness is a spectrum, and it is often difficult to draw a line to separate fair clauses from unfair ones. On the other hand, concrete answers to specific questions leave less room for interpretation.

Guha et al.~\cite{NEURIPS2023_89e44582} presents a large dataset titled ``LegalBench'' consisting of various legal tasks, including detection of unfair terms of service and question answering regarding privacy policy. However, the privacy policy and terms of service documents included in these tasks are very short in length, which is typically not the case in the real-world.

\section{Dataset}\label{sec:dataset}

Our dataset, named \textit{ToS-Busters}, is made of 12 questions applied to the privacy policies of 220 popular online services such as Google and Wikipedia, resulting in a total of 2640 instruction-answer pairs. Ground truth answers for the questions are taken from PrivacySpy\footnote{\url{https://privacyspy.org/}}, which is an open project by a non-profit organization with the mission of grading and monitoring privacy policies for convenience and accountability, licensed under Creative Commons BY. The template used for instructions in our dataset is:

\textit{``In the following privacy policy: [policy], [question1..12]. Answer only with a single number chosen from the following options: [option1..k].''}

The 12 questions are listed below:
\begin{enumerate}
    \item Does the policy allow personally-targeted or behavioral marketing?
    \item Does the policy outline the service's general security practices?
    \item Does the service collect personal data from third parties?
    \item Is the policy's history made available?
    \item Does the service allow you to permanently delete your personal data?
    \item Does the policy require users to be notified in case of a data breach?
    \item Does the service allow third-party access to private personal data?
    \item Is it clear why the service collects the personal data that it does?
    \item Does the service allow the user to control whether personal data is collected or used for non-critical purposes?
    \item When does the policy allow law enforcement access to personal data?
    \item Does the policy list the personal data it collects?
    \item Will affected users be notified when the policy is meaningfully changed?
\end{enumerate}

The corresponding alternatives for each question are listed in Appendix~\ref{appendix}. Table \ref{tab:dataset_compare} compares our dataset with existing datasets in the literature. Note that our dataset contains over four times more terms of service documents compared to CLAUDETTE~\cite{Lippi2019}, and the documents in our dataset are around 50 times longer than LegalBench~\cite{NEURIPS2023_89e44582}. Moreover, unlike CLAUDETTE, our benchmark and baseline investigate the entire document at once, not individual clauses.

\begin{table}[h]
\centering
\caption{Comparison of our dataset with other dataset in the literature}
\resizebox{\linewidth}{!}{
\begin{tabular}{lllll}
\toprule
Name & Documents & Task & Mean Length (Words) \\
\midrule
CLAUDETTE~\cite{Lippi2019} & 50 & Classification & 6,158\\
LegalBench~\cite{NEURIPS2023_89e44582} (privacy\_policy\_entailment task) & 4,343 & Question Answering & 112\\
LegalBench~\cite{NEURIPS2023_89e44582} (privacy\_policy\_qa task) & 10,931 & Question Answering & 41\\
LegalBench~\cite{NEURIPS2023_89e44582} (unfair\_tos task) & 3,822 & Question Answering & 34\\
ToS-Busters (Ours) & 220 & Question Answering & 4,955\\
\bottomrule
\end{tabular}
}
\label{tab:dataset_compare}
\end{table}

\section{Method}\label{sec:method}

\subsection{Question answering}

In contrast to Lippi et al.~\cite{Lippi2019}, that classifies individual clauses using Support Vector Machines (SVMs), our method answers questions regarding the entire document using text generation. 
Text generation is an NLP task where the goal is to generate novel text given another text as input. 
Text generation is a broad task that encompasses various sub-tasks such as text completion and machine translation via text-to-text generation. 
In particular, we use completion generation models, where given an incomplete text, a longer text is generated word-by-word. We use popular pre-trained open source and commercial GPT-based models. Each model is trained using a specific text template, which needs to be followed during inference to get optimal responses.

\subsection{Text summarization}
The lengths of the terms of service documents vary between $62$-$41510$ words, as well as the maximum number of tokens ($t_{max}$) supported by different chatbots.
As a result in some cases, the text needed to be summarized to fit the maximum number of tokens permitted for each bot.

For open-source alternatives, initially the maximum length ($l_{err}$) which generates an error from the chatbot is evaluated empirically by asking questions one by one and recording said maximum.
Thereafter, for every question, the length of which exceeds $l_{err}$, a summarization procedure is carried out as shown in Algorithm~\ref{alg:summary}, where $n$ refers to the number of pieces the input text is split into, $c_{max}$ is the maximum number of characters that can be used to summarize each piece. Initially, the text is split in two pieces, each piece is introduced to the chatbot which is asked to provide a summary with $c_{max}$ characters.
Should the summary of any piece fail, the number of splits is incremented by one and the procedure continues from line $5$.
Otherwise, the summarized pieces are concatenated (represented as addition in line $10$).
The output is sent again to the chatbot to validate whether it is accepted.
If so, the procedure ends. Otherwise $n$ is incremented by one and the procedure continues from line $5$.
The process is repeated until an acceptable summary is produced.
\begin{algorithm}
\caption{Summarization Procedure}
\begin{algorithmic}[1]
\STATE {$n\gets 2$} 
\STATE {valid $\gets False$}
\WHILE{\NOT valid}
\STATE split text in $n$ pieces
\STATE $c_{max}\gets l_{err}/n$
\FOR{i $<$ n } 
\STATE {ask bot to summarize piece $i$ ($p_i$) with $c_{max}$ characters} 
\STATE {$i\gets i +1$} 
\ENDFOR
\STATE {summary = $p_1$ + ... + $p_n$} 
\IF {summary accepted by bot} 
  \STATE valid $\gets True$
\ELSE
  \STATE $n\gets n+1$
\ENDIF 
\ENDWHILE
\end{algorithmic}\label{alg:summary}
\end{algorithm}

For commercial chatbots, the maximum number of allowed tokens is higher, with summarization being needed in only some of the cases.
Additionally, a slightly different process is followed.
The number of tokens needed for each text input is calculated by firstly using the \texttt{encode} function from the \texttt{tiktoken} Python library~\footnote{\url{https://github.com/openai/tiktoken}}, to tokenize said input text (of type string).
Thereafter, the tokens are counted using the \texttt{len()} function.
The required number of splits $n$ is calculated given $t_{max}$ and the number of tokens needed for the input text, such that the number of tokens for each piece is smaller than $t_{max}$. 
This is necessary for the chatbot to be able to accept the piece to be summarized.
As the number of maximum characters per piece cannot be calculated directly, an approximation is used where $c_{max}$ is calculated by dividing the length of the input text by the number of splits.
The chatbots are thereafter queried to summarize the text with less than the evaluated $c_{max}$. 
The merit of this approach is that it is not needed to know $c_{max}$ beforehand.

\section{Experiments}\label{sec:experiments} 
In this section we provide the details of our experimental setup, the chatbots used, how the input was formatted, and the results considering efficiency over all questions and for each question separately.
\subsection{Setup}
Experiments were run for five different chatbots, four of which are open-source, and two commercial ones, specifically chatGPT-3.5-turbo-0125\footnote{\url{https://platform.openai.com/docs/models/gpt-3-5-turbo}}, and chatGPT-4-turbo\footnote{\url{https://platform.openai.com/docs/models/gpt-4-turbo-and-gpt-4}}.
The four selected open-source chatbots are: Nous-Hermes-2-SOLAR-10.7B\footnote{\url{https://huggingface.co/NousResearch/Nous-Hermes-2-SOLAR-10.7B}}, Nous-Hermes-Llama2-13b\footnote{\url{https://huggingface.co/NousResearch/Nous-Hermes-Llama2-13b}}, Mixtral-8x7B-v0.1-Instruct\footnote{\url{https://huggingface.co/mistralai/Mixtral-8x7B-Instruct-v0.1}}, and Smaug-34B-v0.1\footnote{\url{https://huggingface.co/abacusai/Smaug-34B-v0.1}}.
These were chosen as among the top performing chatbots in the Open LLM Leaderboard\footnote{\url{https://huggingface.co/spaces/HuggingFaceH4/open_llm_leaderboard}}.

The token limit for the open-source chatbots is $2048$ or less. In some of the experiments we experienced, for the same chatbot, the limit would go back and forth between $1024$ and $2048$.
A notable exception is the Nous-Hermes-2-SOLAR-10.7B, which required no text summarization.
Regarding the chatGPT versions, the 3.5-turbo-0125 has a token limit of $16385$, whereas the 4-turbo has a limit of $128000$.
These values refer only to the size of the input to the chatbots.
All chatbots are deployed in a Docker container on a server with 2$\times$Nvidia A6000 GPUs, with 48 GBs of memory each.

The same prompt was formatted in three different ways to accommodate for the requirements of the different chatbots.
Specifically, for Nous Hermes alternatives the following was adopted:
\begin{verbatim}
<|im_start|> system
You are a helpful chatbot.
<|im_end|>
<|im_start|>user
[instruction]
<|im_end|>
<|im_start|>assistant
\end{verbatim}

For the rest of the open-source chatbots, the prompt was formatted as follows:
\begin{verbatim}
### Instruction:
[instruction]
### Response:
\end{verbatim}

For the chatGPT alternatives, the prompt was:

\begin{verbatim}
messages=[
    {"role": "system", "content": "You are a helpful assistant."},
    {"role": "user", "content": [instruction]]}
    ]
\end{verbatim}

Each experiment consisted of querying a chatbot for each question, for a total of $220 \times 12$ questions, where $220$ is the number of individual privacy policy documents, and $12$ is the number of distinct questions applied to each privacy policy document. 
The answer of the chatbot was recorded as is.

The accuracy ($\rho$) of each chatbot is calculated per question over all privacy policy documents as: 
\begin{equation}
    \rho_i=\frac{\mu_i}{\delta_i-\lambda_i}\cdot100\%,
\end{equation}
where $\mu_i$ is the number of times the answer to $\rho_i$ matches the ground truth, $\delta_i$ is the total number of times $\rho_i$ has been asked (over the different privacy policy documents), and $\lambda_i$ is the number of times the response of the chatbot resulted in a validation error due the number of tokens being exceeded. 
Although the summarization technique was adopted, there were still few cases where the interaction with the chatbot was not successful. These cases are flagged, thereafter discarded in the calculation of the metrics.
Moreover, the cases can differ between chatbots.

Accuracy is also expressed over all privacy policy documents ($\xi$) as:
\begin{equation}
    \xi = \frac{\sum_{i=1}^{d} \beta_i\cdot q}{(d-\lambda)}\cdot100\%,
\end{equation}
where $d$ is the total number of privacy policy documents ($220$), $q$ is the total number of distinct questions ($12$), skipped is the number $\lambda$ of privacy policy documents discarded due to validation errors from the corresponding chatbot, and $\beta_i$ is the the number of distinct questions answered correctly for privacy policy document $i$.

These metrics are also calculated if the answer to the questions would be chosen randomly among the alternatives. 
Note, different questions have a different number of alternatives.
The random probability per question is calculated as:
\begin{equation}
    \rho_{rand_i} = \frac{1}{\alpha_i}\cdot100\%,
\end{equation}
where $\rho_{rand_i} $ is the probability of answering correctly to question $i$, and $\alpha_i$ is the number of alternative answers for question $i$ from the ground truth.

The overall average accuracy is calculated as:
\begin{equation}
    \xi_{rand} = \frac{\sum_{i=1}^{q} \rho_{rand_i}}{q}\cdot100\%.
\end{equation}

\pgfplotstableread{
Chatbot  Params  Skip  Avg.   Q1  Q2   Q3   Q4   Q5  Q6  Q7  Q8   Q9   Q10  Q11  Q12   

Random  - 0   24.0 25   16.7 33.3 33.3 25    25     20    20    20    16.7 20    33.3 

NH-S   10.7B  0  37.3  58.2  14.1  60.5  28.6  49.6  23.2  28.2  25.5  31.4  16.8  50.9  60.9 

NH-L2    13B  1    32.4  48.4  19.2  52.1  29.7  42  15.5  20.6  21.9  21.9  26.9  29.7  60.7 

Mixtral  56B$^*$  13    44.2  66  34.3  61.3  49.5  49.5  52.2  24.4  18  42.3  15.8  59.3  53.8 

Smaug    34B  8   38.7  63.1  15.9  57  34.1  36.9  15.9  21.5  58.4  41.5  15.9  38.8  61.7 

GPT-3.5      ?$^\dagger$  0   38.0 50.9 23.6 53.2 63.6 41.4 20.9 19.5 38.6 23.2 24.1 32.7 64.1 

GPT-4      ?$^\dagger$    0     53.3 64.1 35.9 69.1 71.4 50.0 64.1 40.0 64.1 38.6 20.9 50.0 71.4 
}\mytable

\pgfplotstabletranspose[string type,
    colnames from=Chatbot,
    input colnames to=Chatbot
]\mytablenew{\mytable}

\begin{table}
\centering
\caption{Total and per question accuracy for the tested chatbots.}\label{tab:results}

\pgfplotstabletypeset[
    ,
    columns/Chatbot/.style={
        string replace={Avg.}{$\xi$},
        string replace={Q1}{$\rho_1$},
        string replace={Q2}{$\rho_2$},
        string replace={Q3}{$\rho_3$},
        string replace={Q4}{$\rho_4$},
        string replace={Q5}{$\rho_5$},
        string replace={Q6}{$\rho_6$},
        string replace={Q7}{$\rho_7$},
        string replace={Q8}{$\rho_8$},
        string replace={Q9}{$\rho_9$},
        string replace={Q10}{$\rho_{10}$},
        string replace={Q11}{$\rho_{11}$},
        string replace={Q12}{$\rho_{12}$},
    },
    every head row/.style={
        before row=\toprule,
        after row=\midrule
    },
    every last row/.style={
        after row={\bottomrule \\
            \multicolumn{8}{l}{$^*$Only 12B parameters are active during inference}\\
            \multicolumn{8}{l}{$^\dagger$Commercial models do not publicly report the number of parameters, and}\\
            \multicolumn{8}{l}{estimates greatly vary}\\
            }
    },
    string type]{\mytablenew}
\end{table}

\subsection{Results}
The results of the experiments are summarized in Table~\ref{tab:results}, showing that all chatbots perform better than the random strategy in selecting the correct asnwer.
The best average accuracy is achieved by chatGPT-4-turbo (gpt-4) with $53.3\%$.
The best average accuracy among open-source chatbots is achieved by Mixtral-8x7B-v0.1-Instruct (referred to as Mixtral in the table) with $44.20\%$ , followed by Smaug-34B-v0.1 (Smaug) with $38.68\%$.
These two chatbots perform at the same level or better as compared to chatGPT-3.5-turbo-0125 (gpt-3.5).
Additionally, it can be noted that, even the best performing chatbot does not perform equally well on all questions.
For example, chatGPT-4-turbo has only $35.9\%$ accuracy for question $2$, and accuracy as high as $71.4\%$ for questions $4$ and $12$.
Similarly, Mixtral-8x7B-v0.1-Instruct, while being the best among open-source and chatGPT-3.5-turbo-0125 on average, the results for questions $8$ and $10$, are quite low, i.e., $18.01\%$ and $15.79\%$ respectively, performing worse than random.
The lowest performance is recorded for question $10$, where Nous-Hermes-Llama2-13b (NH-L2), chatGPT-3.5-turbo-0125 (gpt-3.5), and chatGPT-4-turbo (gpt-4) report the best accuracies.
It is not known at the time of writing why some questions are more difficult to answer than others.

We manually extracted the privacy policy documents that resulted in validation errors across chatbots, in this case specifically when using Nous-Hermes-Llama2-13b, Mixtral-8x7B-v0.1-Instruct, and Smaug-34B-v0.1.
Some overlap was noted, i.e., the same document resulted in a validation error in all three of the aforementioned models, while some were problematic only in one or two.
All the questions related to any of these privacy policy documents 
were removed from consideration in the calculation of the performance metrics for each chatbot.
These results are shown in Table~\ref{tab:results-filtered}.
Note the random column has no change, and has only been added to aid the reader.

\pgfplotstableread{
Chatbot  Params  Avg.   Q1  Q2   Q3   Q4   Q5  Q6  Q7  Q8   Q9   Q10  Q11  Q12   
Random  -    24.0 25   16.7 33.3 33.3 25    25     20    20    20    16.7 20    33.3 

NH-S          10.7B   37.6  60.0 14.6 60.5 28.3 49.8 23.4 27.3 25.4 31.7 18.0 51.2 60.97 

NH-L2   13B   32.3  48.3 20.0 51.7 29.8 41.9 14.6 20.5 21.5 22.4 26.8 28.8 60.5

Mixtral  56B$^*$   43.8  65.8 34.6 61.5 48.3 50.2 52.2 24.4 18.0 42.4 15.6 59.5 53.2 

Smaug            34B   38.4  64.9 16.1 57.1 34.6 36.6 14.6 21.5 57.6 41.9 16.1 39.0 61.5 

GPT-3.5      ?$^\dagger$   38.2  52.2 24.9 53.2 63.9 41.5 20.97 20.0 38.0 23.4 24.88 31.7 63.9 

GPT-4      ?$^\dagger$            53.8  64.9 35.6 68.3 71.2 49.8 65.4 41.0 64.9 40.5 22.0 51.2 70.7  
}\mytable

\pgfplotstabletranspose[string type,
    colnames from=Chatbot,
    input colnames to=Chatbot
]\mytablenew{\mytable}

\begin{table}
\centering
\caption{Total and per question accuracy for the tested chatbots excluding datasets skipped by at least one chatbot.}\label{tab:results-filtered}

\pgfplotstabletypeset[
    ,
    columns/Chatbot/.style={
        string replace={Avg.}{$\xi$},
        string replace={Q1}{$\rho_1$},
        string replace={Q2}{$\rho_2$},
        string replace={Q3}{$\rho_3$},
        string replace={Q4}{$\rho_4$},
        string replace={Q5}{$\rho_5$},
        string replace={Q6}{$\rho_6$},
        string replace={Q7}{$\rho_7$},
        string replace={Q8}{$\rho_8$},
        string replace={Q9}{$\rho_9$},
        string replace={Q10}{$\rho_{10}$},
        string replace={Q11}{$\rho_{11}$},
        string replace={Q12}{$\rho_{12}$},
    },
    every head row/.style={
        before row=\toprule,
        after row=\midrule
    },
    every last row/.style={
        after row={\bottomrule \\
            \multicolumn{8}{l}{$^*$Only 12B parameters are active during inference}\\
            \multicolumn{8}{l}{$^\dagger$Commercial models do not publicly report the number of parameters,}\\
            \multicolumn{8}{l}{and estimates greatly vary}\\
            }
    },
    string type]{\mytablenew}
\end{table}

Additional tests were conducted to investigate whether the chatbots are able to also provide the line in the inputted text from which the answer is taken/inferred from.
In this case only Mixtral-8x7B-v0.1-Instruct was used for all privacy policies and questions as it is the top performing chatbot among the open-source alternatives.
A lower performance was observed. The overall accuracy is $38.11\%$, and for each question from Q1-12 the values for accuracy are respectively $50.95\%$, $39.23\%$, $48.58\%$, $43.81\%$, $45.71\%$, $39.23\%$, $20.57\%$, $12.02\%$, $39.13\%$, $17.31\% $,$47.85\%$, $50.47\%$.
It is not known why the performance drops when querying for the line the document as well. 
It is not clear how the components of the task are prioritized, that is, whether the chatbot focuses on the first part about answering the question, or the second part regarding the extraction of the line that contains the answer.
A line from the privacy policy is provided, even in those cases when the answer is incorrect.

\section{Conclusive Remarks}\label{sec:conclusion} 
In this paper we evaluated the efficacy of LLMs, both open-source and commercial, in answering questions about long privacy policies from well-known websites.
While both types of chatbots perform better than a random approach would, there is room for improvement, especially when dealing with long text documents.
The latter is true also for the best performing chatbot investigated in this paper, that is chatGPT-4-turbo.
Most importantly, we introduce a dataset with $12$ questions asked to a set of privacy policies.
The questions and corresponding answers are intended to uncover or emphasize what potentially could be dubious clauses, to which online users agree to.
Indeed, most users sign these online contracts without much deliberation, aiming to get a desired service as fast as possible, having potentially signed their rights away with a few clicks.
It is clear that such (online) interaction with different companies and institutions poses grave threats to people's privacy, and in some cases, also to body autonomy itself.
While the authors agree with Vardi~\cite{vardi2023}, that such contracts should be nullified, the road to that legislation seems to be quite long.
In the meantime, online users are in need of assistance in their daily online activities.

An interesting future research direction is finding ways to improve LLMs for the task evaluated in this paper, so that they can be adopted widely by online users, aiding them to avoid predatory click-through contracts.
Another interesting research direction is studying the effects of summarization, as well as the variance of prompts in the performance of LLMs, and their optimization thereof.


\bibliographystyle{unsrt}
\bibliography{biblio}

\appendix

\section{List of Dataset Questions}\label{appendix}

The $12$ questions, and their alternative responses, used in the experiments are:

\begin{enumerate}
    \item Does the policy allow personally-targeted or behavioral marketing?
    \begin{itemize}
      \item Yes
      \item Yes, but you may opt-out
      \item Yes, but you must opt-in
      \item No
    \end{itemize}
    \item Does the policy outline the service's general security practices?
    \begin{itemize}
        \item No
        \item Somewhat
        \item Yes
        \item Yes, including audits
        \item Yes, including independent audits
        \item N/A
    \end{itemize}
    \item Does the service collect personal data from third parties?
    \begin{itemize}
        \item Yes
        \item Yes, but for critical data only
        \item No
    \end{itemize}
    \item Is the policy's history made available?
    \begin{itemize}
        \item No
        \item Only the date it was last modified
        \item Yes, with revisions or a changelog
    \end{itemize}
    \item Does the service allow you to permanently delete your personal data?
    \begin{itemize}
        \item No
        \item Yes, by contacting someone
        \item Yes, using an automated mechanism
        \item N/A
    \end{itemize}
    \item Does the policy require users to be notified in case of a data breach?
    \begin{itemize}
        \item Not necessarily
        \item Yes, eventually
        \item Yes, within 72 hours
        \item N/A
    \end{itemize}
    \item Does the service allow third-party access to private personal data?
    \begin{itemize}
        \item Yes
        \item Yes, all parties specified (including non-critical service providers such as advertisers)
        \item Yes, not all parties specified (but only to critical service providers)
        \item Yes, all parties specified (only to critical service providers)
        \item No
    \end{itemize}
    \item Is it clear why the service collects the personal data that it does?
    \begin{itemize}
        \item No
        \item Somewhat
        \item Mostly
        \item Yes
        \item N/A
    \end{itemize}
    \item Does the service allow the user to control whether personal data is collected or used for non-critical purposes?
    \begin{itemize}
        \item No
        \item On an opt-out basis, but only for some non-critical data/uses
        \item On an opt-out basis, for all non-critical data/uses
        \item On an opt-in basis
        \item N/A (no data used for non-critical purposes)
    \end{itemize}
    \item When does the policy allow law enforcement access to personal data?
    \begin{itemize}
        \item Always
        \item Not specified
        \item When reasonably requested
        \item Only when required by a court order or subpoena
        \item N/A (no personal data to share)
        \item Never (special legal jurisdiction)
    \end{itemize}
    \item Does the policy list the personal data it collects?
    \begin{itemize}
        \item No
        \item Only summarily
        \item Yes, generally
        \item Yes, exhaustively
        \item N/A (no personal data is collected)
    \end{itemize}
    \item Will affected users be notified when the policy is meaningfully changed?
    \begin{itemize}
        \item No
        \item Yes
        \item N/A
    \end{itemize}
\end{enumerate}

\end{document}